\documentclass[letterpaper, 10 pt, conference]{ieeeconf}
\IEEEoverridecommandlockouts
\usepackage{cite}
\usepackage{amsmath,amssymb,amsfonts}
\usepackage{algorithmic}
\usepackage{graphicx}
\usepackage{textcomp}
\usepackage{xcolor}
\usepackage{gensymb}
\usepackage{siunitx}
\usepackage{dsfont}
\usepackage{pgfplots}
\pgfplotsset{compat=1.18}

\usepackage[ruled,vlined,linesnumbered]{algorithm2e}
\SetAlCapNameFnt{\scriptsize}
\SetAlCapFnt{\scriptsize}

\def\BibTeX{{\rm B\kern-.05em{\sc i\kern-.025em b}\kern-.08em
    T\kern-.1667em\lower.7ex\hbox{E}\kern-.125emX}}

\title{Learning the Approach During the Short-loading Cycle Using Reinforcement Learning}

\author{Carl Borngrund$^{1}$ and Ulf Bodin$^{2}$ and Henrik Andreasson$^{3}$ and Fredrik Sandin$^{4}$
\thanks{This research was conducted with support from Sweden's Innovation Agency and the VALD project under grant agreement no. 2021-05035.}
\thanks{$^{1}$ Carl Borngrund is with EISLAB, Luleå University of Technology, 97187 Luleå, Sweden {\tt\small carl.borngrund@ltu.se}}%
\thanks{$^{2}$ Ulf Bodin is with EISLAB, Luleå University of Technology, 97187 Luleå, Sweden {\tt\small ulf.bodin@ltu.se}}
\thanks{$^{3}$ Henrik Andreasson is with the Centre for Applied Autonomous Sensor Systems, Örebro University, 70182, Sweden {\tt\small henrik.andreasson@oru.se}}
\thanks{$^{4}$ Fredrik Sandin is with EISLAB, Luleå University of Technology, 97187 Luleå, Sweden {\tt\small fredrik.sandin@ltu.se}}
}

\begin{document}

\maketitle
\thispagestyle{plain}
\pagestyle{plain}

\begin{abstract}
The short-loading cycle is a repetitive task performed in high quantities, making it a great alternative for automation.
In the short-loading cycle, an expert operator navigates towards a pile, fills the bucket with material, navigates to a dump truck, and dumps the material into the tipping body.
The operator has to balance the productivity goal while minimising the fuel usage, to maximise the overall efficiency of the cycle.
In addition, difficult interactions, such as the tyre-to-surface interaction further complicate the cycle.
These types of hard-to-model interactions that can be difficult to address with rule-based systems, together with the efficiency requirements, motivate us to examine the potential of data-driven approaches.
In this paper, the possibility of teaching an agent through reinforcement learning to approach a dump truck's tipping body and get in position to dump material in the tipping body is examined.
The agent is trained in a 3D simulated environment to perform a simplified navigation task.
The trained agent is directly transferred to a real vehicle, to perform the same task, with no additional training.
The results indicate that the agent can successfully learn to navigate towards the dump truck with a limited amount of control signals in simulation and when transferred to a real vehicle, exhibits the correct behaviour.
\end{abstract}

\section{Introduction}
Much of the labour within the construction industry requires heavy-duty machinery which is explicitly designed to execute construction tasks, often including earthworks \cite{10.22260/ISARC2011/0217}.
Examples of such heavy-duty machinery are wheel loaders and dump trucks, where wheel loaders are used to scoop some material and dump it nearby.
In contrast, dump trucks are used to transport larger amounts of material longer distances.

\begin{figure}
    \centering
    \includegraphics[width=\columnwidth]{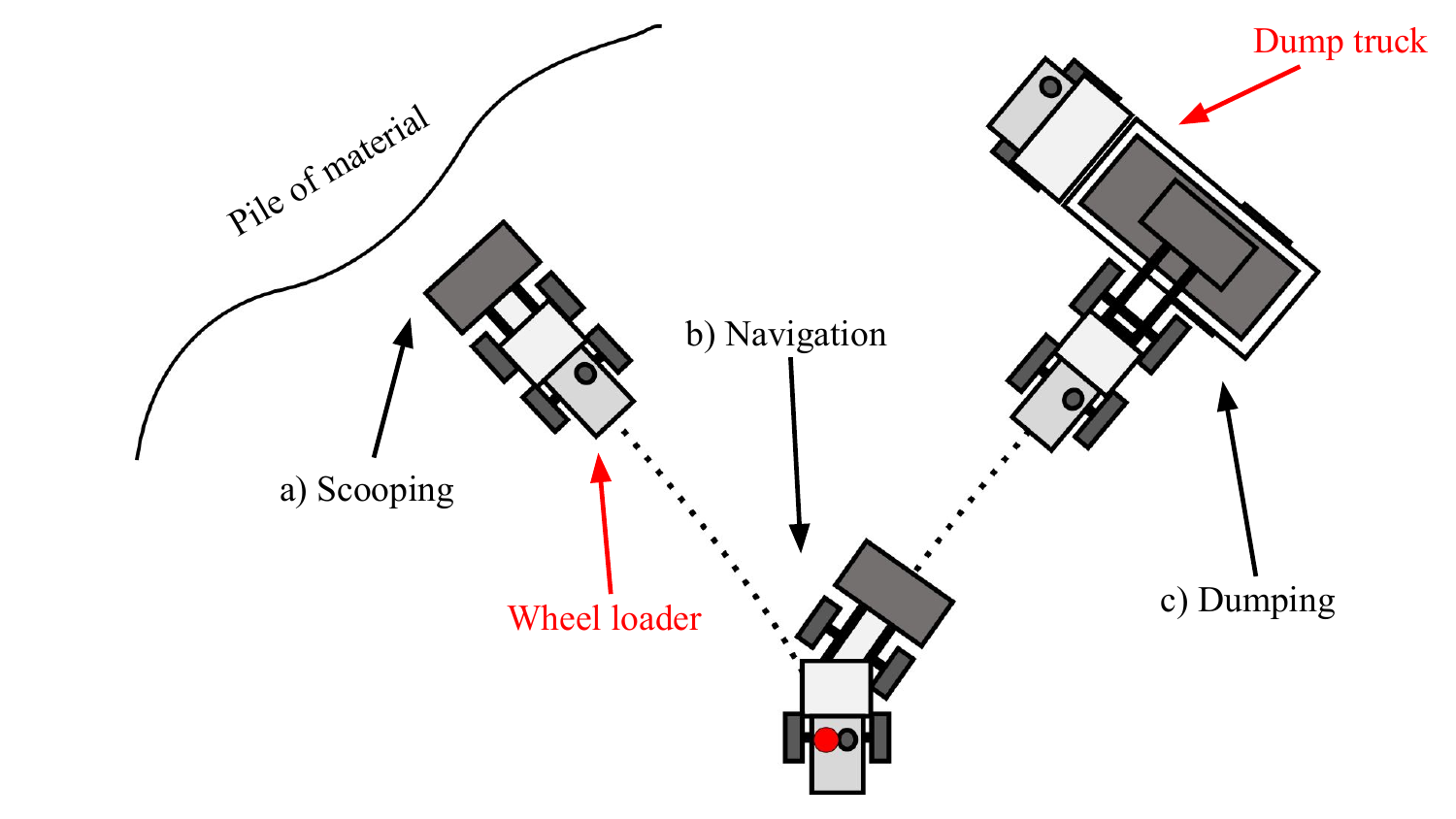}
    \caption{This is an overview of the short-loading cycle, where the objective is for the operator of the wheel loader to transfer material from a pile to the tipping body of a dump truck. The short-loading cycle consists of three tasks required for its completion: a) scooping, b) navigation, and c) dumping.}
    \label{fig:slc}
\end{figure}

One of the tasks within the construction industry is the short-loading cycle, which is a cooperative task between the wheel loader operator and a dump truck operator, as shown in Figure \ref{fig:slc}.
The goal is for the operator of the wheel loader to scoop material from a nearby pile, navigate to the tipping body of the dump truck, and unload the scooped material into the tipping body.

The short-loading cycle is a highly repetitive task performed in high quantities making it a great candidate for automation.
The short-loading cycle is often part of a large refinement process, such as open pit mining, where any large variance in performance will affect the full process.
Automation can alleviate the variances in performance between operators \cite{0381b348-1045-4d71-b336-50a61b80208a}.

The short-loading cycle can be divided into three main tasks: scooping, navigation, and dumping.
Navigation has previously been addressed mostly by rule-based systems \cite{SHI2020103570, https://doi.org/10.1111/0885-9507.00222, 4650638}, with only a few works attempting to leverage deep-learning-based solutions \cite{10260481}.
Deep-learning-based solutions have shown great potential in terms of matching expert operator performance on the bucket-filling task \cite{9206849}.

Previous work mainly occurred in simulated environments or on miniature wheel loaders, with limited attempts at real-world navigation.
Few studies explored the use of deep-learning techniques for automating the navigation of the short-loading cycle.
This, together with the success of deep learning on the bucket-filing problem, motivated an exploration into the possibility of using reinforcement learning (RL) to effectively handle the navigation phase.
RL has the potential to result in a solution that is capable of generalising to unseen environments, being trained without the requirements of operator data, and improving performance online.

\begin{figure*}
    \centering
    \includegraphics[width=\textwidth]{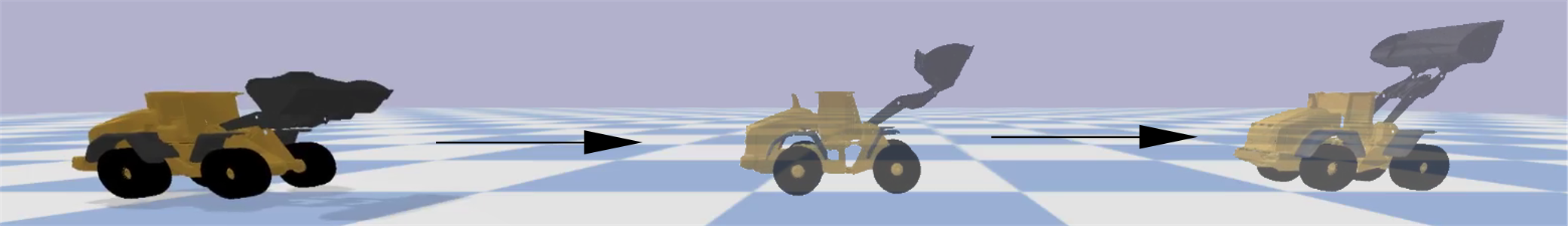}
    \caption{The simulated setup where the agent should learn how to drive forward while lifting the bucket.}
    \label{fig:simulated_env}
\end{figure*}

In this work, an agent is trained in simulation to navigate to a pre-defined point, while lifting the bucket, to imitate approaching a dump truck.
The agent controls brake and lift, while a PID regulates the forward motion of the vehicle.
The agent gets a reward based on correctly navigating towards the point while lifting the bucket to a goal height as fast as possible.
The results indicate that the agent can successfully be trained in simulation to perform the task at high performance.
Figure \ref{fig:simulated_env} shows the simulated setup.
After transfer to a real machine, the agent exhibited the correct behaviour in a real environment which included a time-delay issue in the positioning sensor.
The results seem to indicate that this type of development pipeline has potential, however, due to the limited scope of the experiments more work is required to fully verify it.

The main contributions of this paper relate to:
\begin{itemize}
    \item Formulation of a reward function capable of training an agent to perform the approach towards a dump truck in simulation.
    \item A successful simulation-to-reality transfer of the trained agent performing a simultaneous lift and navigation task.
    \item Reasoning relating the transfer between simulation and reality for further automation of the navigation during the short-loading cycle.
\end{itemize}

The remainder of the paper is structured as follows:
Section \ref{section:related_work_problem_analysis} presents some selected related work together with a problem analysis discussing the challenges related to automating the navigation of the short-loading cycle.
Section \ref{section:background} presents some background information concerning the short-loading cycle and RL.
Section \ref{section:experimental_setup} describes the experimental setup, how the wheel loader was modelled, the simulated environment, how the agent was trained, and how the agent was validated on a real wheel loader.
Section \ref{section:results} presents the results from the training of the agent in the simulated environment and the validation on the real wheel loader, which are discussed in Section \ref{section:discussion}.
Lastly, the paper is concluded in Section \ref{section:conclusion}, where some possible directions for future work are presented.

\section{Related Work \& Problem Analysis} \label{section:related_work_problem_analysis}
The short-loading cycle is a repetitive task performed in high quantities and is often part of a larger process in, for example, an open pit mine, where the pile is continuously being refilled with new raw material that needs to be moved to another location for refinement.
This means that any inefficiencies or unplanned stops will have a cascading effect on the downstream task.
The efficiency of the cycle can have a large variance depending on the operator skill \cite{0381b348-1045-4d71-b336-50a61b80208a}.
The operating cost can loosely be divided into 3 equal parts: operator salary, fuel costs, and maintenance costs.
These reasons together make the short-loading cycle a suitable candidate for automation as a high-performing autonomous system can hopefully maximize vehicle utilization in terms of fuel and maintenance costs while matching the operator's productivity.
The realisation of an autonomous system can help alleviate some of these large variances, leading to a more reliable performance output \cite{bobbie_shadow_driver}.

On a high level, the short-loading cycle can be divided into three tasks: scooping, navigation, and dumping.
In this work, the main focus is on the navigation task.

A large amount of rule-based approaches have been applied to the navigation problem during the short-loading cycle.
These include clothoid-based solutions for trajectory generation and tracking \cite{https://doi.org/10.1111/0885-9507.00222}, genetic algorithm-based trajectory generation \cite{6704123}, Redd and Shepps for trajectory generation using a PID for tracking \cite{ALSHAER20135315}, RRT* and CC-steer to plan the trajectory and APC for tracking \cite{SHI2020103570}, or performing offline trajectory planning using a non-linear MPC and online trajectory tracking using LVP-MPC \cite{9867662}.
The clothoid-based solution was also used to perform the full cycle \cite{4650638} at a low speed where a single round took around 60 seconds.

Little work has been done using deep learning techniques to address navigation during the short-loading cycle.
The optimal switching time of changing from backwards momentum and forward momentum was examined by approximate dynamic programming together with a neural network \cite{10.1115/DSCC2019-9248}.
They found that the optimal switching time was 2.86 seconds.
RL has been used to perform navigation in simulation with a large reality gap \cite{10260481}.

Deep learning techniques have shown good potential when it comes to filling the bucket with material.
Imitation learning has been used to perform the bucket-filling action on real vehicles, matching the performance of an expert operator \cite{DADHICH20191}.
This solution was then adapted using RL to learn how to scoop another type of material \cite{9206849}.
RL has also been used to navigate to the scooping point and scoop material in high-fidelity simulation \cite{machines9100216}.
RL has shown potential in terms of simulation-to-reality transfer by learning how to scoop material in a simulated world from where the trained agent could successfully be transferred to a miniature wheel loader and scoop material, with no additional training \cite{9344588}.

Apart from the general motivation of automating the short-loading cycle above, the navigation of the short-loading cycle presents us with challenges. 
These challenges include complex and difficult-to-model interactions, such as the terrain-tyre interaction where the load distribution on the wheels \cite{10.1007/978-3-030-54814-8_17}, together with terrain type, terrain conditions, and tyre thread \cite{metz1993dynamics} affect the behaviour of the wheel loader.
The environment is highly dynamic where choices made throughout a round of the cycle, will change how the next round is performed.
This includes but is not limited to scooping position, navigation path, and dumping position.

In addition to the challenges outlined above, a solution based on RL has the potential to adapt and generalize to a diverse set of environments with minimal engineering effort, further improve the performance online, and address the gap between training and deployment environments.
Furthermore, as RL does not leverage expert data to learn, this type of solution can learn to perform tasks on machines where human data is not available, such as wheel loaders created for automation \cite{volvo_lx03}.
The automation challenges facing the short-loading cycle and the strengths of RL described above, together with the successful attempts for other parts of the short-loading cycle, motivated us to examine the possibility of employing RL for navigation.

RL is currently difficult to train from randomly initiated directly in the target environment due to sample inefficiency, stability during training, and the large data requirement.
These challenges lead to the demand to train the agent in some proxy environment and then transfer the trained agent to the target environment.
The two options presented in the literature are to either train in simulation or train on a miniature wheel loader.
Miniature wheel loaders suffer from problems similar to those described above, making the only plausible approach to train in a simulated environment.
This means that the modelling challenges outlined above are still relevant, making the model required to be able to bridge the simulation-to-reality gap.

\section{Preliminaries \& theory} \label{section:background}
\subsection{The short-loading cycle}
As mentioned, the short-loading cycle can, on a high level, be divided into three tasks.
These tasks include navigation, scooping, and dumping \cite{DADHICH2016212}, as shown in Figure \ref{fig:slc}.

To start the short-loading cycle the operator first has to locate from where to scoop in the pile and navigate towards that spot while positioning the bucket level with the ground. 
The operator increases the speed of the vehicle to penetrate the pile properly.

When in the correct position, the operator leverages the throttle, lift, and tilt functionality to fill the bucket with material.
The operator needs to use the weight of the pile to push down the front wheels of the wheel loader to ensure that the wheels do not lose traction.
If done incorrectly the wheels will lose traction and spin rather than provide force, leading to a more difficult bucket-filling phase and generating extra wear on the tyres.
A good strategy for filling the bucket with the material is the so-called "slicing cheese" \cite{filla2017towards} strategy where the operator lifts the boom while simultaneously tilting the bucket to fill the bucket.

Once the bucket has been filled with material the operator reverses from the pile to the reversal point and uses the lift functionality to lift the bucket further into the air.
While reversing from the pile and the reversal point the bucket height usually travels from the height at which it exited the pile to around 50\% of the required lift to get the bucket over the tipping body of the dump truck.

When the lift is sufficiently high, the operator changes direction from reversing to driving forward, to navigate towards the tipping body.
This is done while continuing to lift the bucket upwards.
The selected trajectory for navigation ensures that the operator can reach the dumping position without the need to stop.
During the navigation, the operator needs to be attentive to how the vehicle is moving in terms of the steering actuation.
Due to the weight of the wheel loader will exhibit understeering behaviour requiring extra steering actuation to move in the desired trajectory.
Since the short-loading cycle needs 3-4 rounds to fill a single tipping body the operator needs to dump into either the front, middle, or back of the tipping body to fill it evenly.
Once the bucket is above the dump truck, no large changes can be made where to dump, meaning that this decision has to be made during the navigation.

After, the operator begins to dump the material that's in the bucket into the tipping body of the dump truck.
While dumping, the operator needs to make sure that the material is evenly spread in the tipping body after 3-4 rounds to ensure correct weight distribution for the dump truck.
This is achieved by never being stationary while dumping but rather starting when the vehicle is still moving, although slowly, forward towards the dump truck and reversing away from the dump truck before the bucket is empty, together with small steering actuation.

Once the bucket is emptied, the operator reverses away from the dump truck, while lowering the bucket into the scooping position, getting ready for the next round of the cycle.

\subsection{Reinforcement learning}
Reinforcement learning is a training paradigm within the domain of machine learning where the objective is to find the mapping from observation to action through the use of a reward function which describes how the agent should behave to perform some task.
RL is often used for tasks where it can be difficult to create a dataset due to there being no single mapping from input to output.
As RL is driven by a reward function, it can be advantageous to describe the success of tasks through this function, rather than through data.
For example, the performance of the short-loading cycle is often measured in productivity and fuel efficiency \cite{Filla438934}.
This means that if important metrics can successfully be reflected in a reward function, the agent will optimize towards this and hopefully lead to a high-performing agent.
However, it is non-trivial to translate a sparse reward, such as productivity, into a reward given in each timestep.

Formally, an environment consists of a state space, $\mathcal{S}$, an action space, $\mathcal{A}$, state transition probabilities $\mathcal{P}^a_{ss'} = \mathds{P}[\mathcal{S}_{t+1}=s' | S_t = s, A_t = a]$, and a reward function $\mathcal{R}$.
In each timestep, the agent chooses an action, $a$, based on the current state, $a \sim \pi(a | s)$, where $\pi$ is a network with trainable parameters.
From this, the new state, $s'$ is generated, together with a reward $r$.
The $\pi$ selects $a$ by maximizing $\mathds{E}[R_t | s_t] = \mathds{E}_\pi\Bigr[\sum_{k=0}^\infty \gamma^kR_{t+k+1}\Bigr| S_t = s, A_t = a\Bigr]$, which is the expected discounted reward.
Here, $\gamma$ is the discount factor used to discount the rewards further in the future compared to rewards closer in time to the current action.

Actor-critic is an RL method where the critic learns a policy for the agent to determine the next action, while the critic estimates the expected reward based on the actions which provides feedback to the actor. 
The critic estimates the expected reward using $V = \mathds{E}_\pi[R_t|s_t]$ and from this, the actor updates its policy.
One of the methods under actor-critic is the advantage actor-critic method (A2C), which leverages an advantage function.
The advantage function describes the difference between the estimated reward of taking an action in a state, compared to the reward of taking any action in the state, such that $A_\pi(s_t, a_t) = Q_\pi(s_t, a_t) - V_\pi(s_t)$. 
The default action is given by the policy, $\pi$.
This way of calculating $A_\pi(s_t, a_t)$ will yield the lowest variance to other known options \cite{https://doi.org/10.48550/arxiv.1506.02438}.
Finding the true $A_\pi(s_t, a_t)$ is non-trivial, especially with large action and observation spaces, thus estimating the function through some algorithm such as Generalized Advantage Estimation \cite{https://doi.org/10.48550/arxiv.1506.02438} is essential.

\subsection{Proximal policy optimization}
Proximal policy optimization (PPO) \cite{https://doi.org/10.48550/arxiv.1707.06347} is an on-policy A2C algorithm where clipping of the advantage function is used in the objective function to limit the difference between the current policy and updated policy.
To achieve this, a probability ratio, see Equation \ref{PPO:rt}, is added to the objective function, see Equation \ref{PPO:loss_function}, which is a ratio between the current policy and future policy. 
If the ratio is too large, the policy is clipped in the range of $[1 - \epsilon, 1 + \epsilon]$, creating a surrogate probability ratio that can be used in the policy update.
Updating the policy in this type of way leads to higher stability in the training, minimizing the chance of the training collapsing.

\begin{equation} \label{PPO:rt}
    r_t(\theta) = \dfrac{\pi_\theta(a_t | s_t)}{\pi_{\theta_{old}}(a_t | s_t)}
\end{equation}

\begin{equation} \label{PPO:loss_function}
    L^{clip}(\theta) = \Hat{\mathds{E}}\Bigr[ min(r_t(\theta)\Hat{A_t}, clip(r_t(\theta), 1 - \epsilon, 1 + \epsilon)\Hat{A_t}) \Bigr]
\end{equation}

\section{Experimental Setup} \label{section:experimental_setup}
The environment is modelled using PyBullet \cite{coumans2021}, which is a physics engine implemented in C++ with a Python API, together with OpenAI gym \cite{1606.01540} which is a wrapper that provides the RL loop.

In this work, some delimitations and assumptions were made in terms of the short-loading cycle.
Firstly, the task examined is the navigation from the reversal point to the dumping point, where the dumping point is assumed to be an apriori.
Secondly, the max velocities of joints and the machine are enforced through rules in lower ranges to ensure safe operations.
Lastly, no physical object is put in front of the wheel loader, such as a dump truck and instead only abstract GNSS positions are used.
Further limitations regarding the real machine are explained in Section \ref{section:exp_setup:real_env_setup}.

\begin{table}
\centering
\caption{Hyperparameters used are based on the hyperparameters used in \cite{10260481}.}
\label{table:experimental_setup:hyperparameters}
    \begin{tabular}{c|c}
        Hyperparameter & Value \\
        \hline
        Learning rate & $3\cdot10^{-5}$ \\
        n\_steps & $512$ \\
        batch size & $128$ \\
        n\_epochs & $20$ \\
        gamma & $0.99$ \\
        gae\_lambda & $0.9$ \\
        clip\_range & $0.4$ \\
        ent\_coef & $0.0$ \\
        vf\_coef & $0.5$ \\
        max\_grad\_norm & $0.5$ \\
        sde\_sample\_freq & $4$ \\
        use\_sde & True \\
        n\_envs & 1 \\
    \end{tabular}
\end{table}

\subsection{Wheel loader modelling}
The wheel loader model is modelled based on CAD files of a Volvo 150H wheel loader, from where the visual and collision volumes are generated.
These are then used in Unified Robot Description Format (URDF) \cite{tola2023understanding} which is an XML-based format used to describe, for example, robotic models.
URDF describes the robotic model as a directed acyclic graph (DAG) where each vertice describes some volume of the wheel loader, such as the cab or frame.
The edges in the DAG represent how each volume is connected and how they can move relative to each other.

\begin{figure}
    \centering
    \includegraphics[width=\columnwidth]{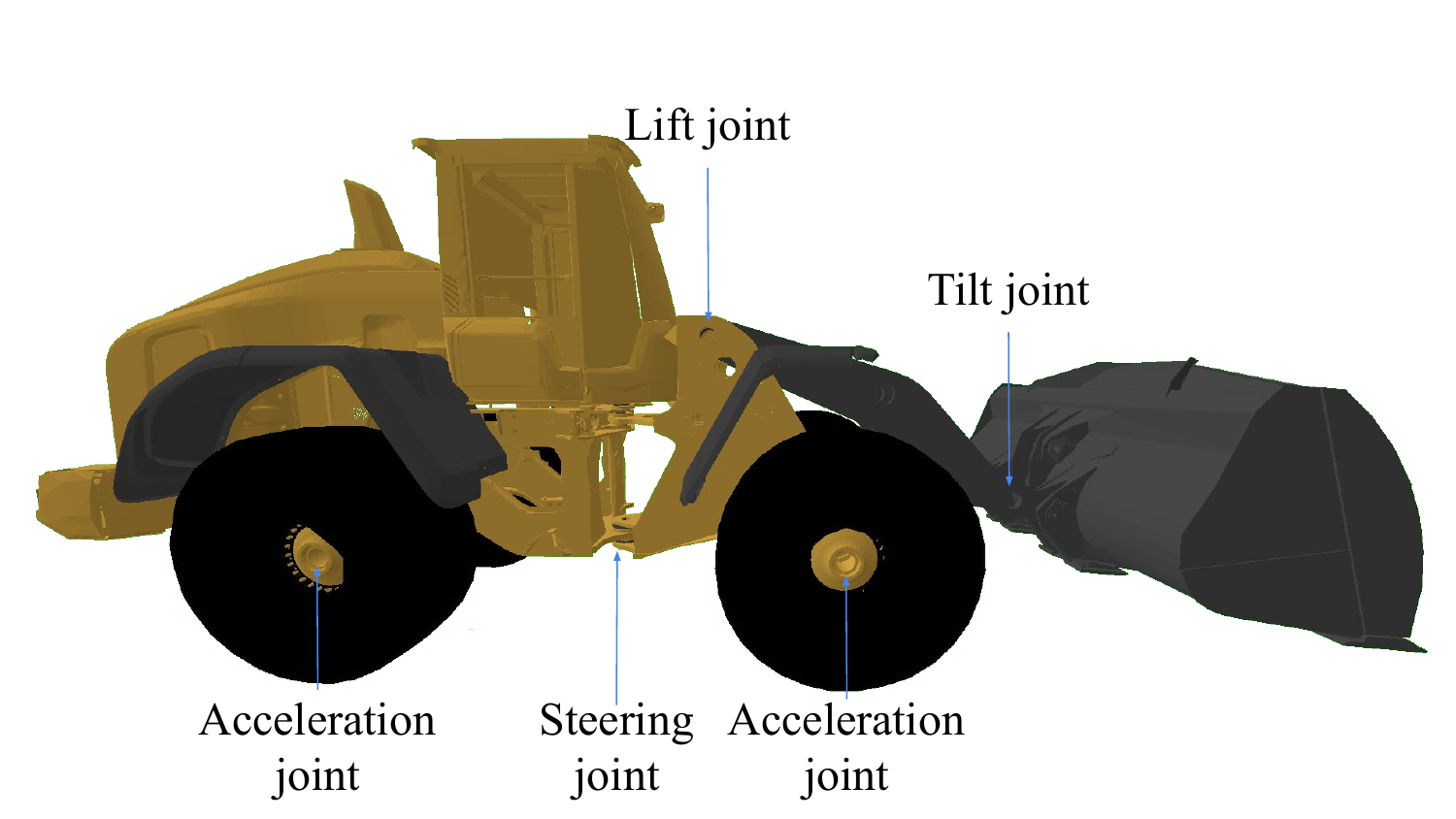}
    \caption{Visualisation of joints of the L150H wheel loader.}
    \label{fig:wheel_loader}
\end{figure}

The wheel loader model, as seen in Figure \ref{fig:wheel_loader}, contains 4 movable joints that are required to control the vehicle.
These movable joints are steering, wheel rotation, lift, and tilt.
The movable joints can be divided into two categories, continuous and revolute joints.
The continuous joints can freely rotate around some axis, such as the wheels being able to rotate to move the vehicle, while the revolute joints are only defined within some limited range.
For example, the steering joint is valid only within the range of [-37.5\degree, 37.5\degree].

The properties of the wheel loader, such as maximum lift height, weight and max velocity are based on the physical information within the L150 datasheet \cite{l150h_datasheet}.
However, some simplifications have been made.
Firstly, the back axle is completely fixed on the model compared to the back axle on a real vehicle capable of tilting.
Secondly, the hydraulics, engine power, or brake was not modelled, and the vehicle is assumed to move at a constant speed if the brake is not engaged.
Lastly, the tyres of the wheel are modelled to be non-deformable.

\subsection{Simulated environment setup}
The simulated environment uses the PyBullet physics engine \cite{coumans2021} for the physics calculations and joint manipulation.
The environment consists of a wheel loader, flat ground with consistent friction and an abstract point to where the agent should navigate.

PyBullet attaches PD regulators to each movable joint of the wheel loader, which was defined in the URDF.
The PD regulators are used to control the motion of each joint, where the joint can either be moved by setting a target velocity, such as for the wheels, or a target position, as for the lift joint.
The joint velocity of each joint was matched between the simulated environment and the real environment.
Due to the weight of the real machine, there are a lot of safety concerns with having an autonomous system perform some tasks, meaning that all joint velocities are limited to relatively low values.
The matching between the simulated and real environment is further described in Section \ref{section:exp_setup:real_env_setup}.

To facilitate the training of the RL agent, the OpenAI gym library \cite{1606.01540} was used, which allows for easy creation of the observation, action, reward, and next observation loop.
The gym library enforces the observation and action structure, to ensure that the agent is never provided with or provides information that is out of bounds or incorrectly typed.

The main task to be learned is for the agent to navigate towards a point, in a straight line, and stop within a certain vicinity with the boom raised above 95\% of the maximum lift.
To achieve this the wheel loader is moving at a constant speed of 2 m/s, and the agent should learn how to control the brake together with the lift to come to a complete stop.
During training, the initial heading of the wheel loader will be randomly chosen and the stopping point was selected at a fixed distance of 5 metres in front of the wheel loader.
This is to achieve variance in the environment, hopefully achieving better resistance to variances in the real world.

The observation defined from the environment consists of the vehicle position relative to the point, the vehicle velocity, and the current lift angle.
From this, the agent should predict two binary values.
The agent should predict if the brake should be engaged or not, and if the boom should move up or not.
For simplicity, the x and y positions are always positive, regardless of heading.

For feedback on whether or not the action predicted from the current observation was good or bad a reward function was used.
The reward function needs to be well-defined in each timestep as the agent will attempt to maximize the expected value of said reward function.
The reward function by which the agent is guided is described in Algorithm \ref{alg:experimental_setup:reward_function}.

\begin{algorithm}
\footnotesize
\caption{Reward function used during training.} \label{alg:experimental_setup:reward_function}
    
$o_r \gets$ flag for outside the valid position range\;
$t_m \gets$ flag for maximum time spent in the episode\;
$done \gets 0$\;
$r \gets 0$\;

\eIf{$\ o_r\ ||\ t_m$}{
    $r \gets -1$\;
    $done \gets 1$\;}
{
    $p_e \gets$ previous distance to end point\;
    $c_e \gets$ current distance to end point\;
    $p_l \gets$ previous lift position\;
    $c_l \gets$ current lift position\;
    $v \gets$ velocity of the wheel loader\;
    $m_l \gets$ max lift height\;
    $t_a \gets$ accumulated timestep\;
    $t_c \gets$ time punishment constant\;
 
    \eIf{$c_e < 1.5$ \& $v < 0.1$ \& $c_l > 0.95 m_l$}{
        $done \gets 1$\;
        $r \gets 1$\;
        }
    {
        $r \gets (p_e - c_e) + (p_l - 0.95 m_l c_l) - t_c t_a$\;
    }
}
return $r$, $done$
\end{algorithm}

The proposed reward function is divided into three different blocks.
The first blocks (lines 5 to 7) describe the negative termination states where the episode terminates if the vehicle is too far away from the starting position or if the maximum time limit is reached.
The second block (lines 17 to 19) describes the positive termination states where the agent has finished the task if the vehicle is close to the endpoint with a low speed and the boom has been lifted to 95\% of the max height.
The last block (line 21) describes the non-terminal reward where a reward is given based on whether the agent is moving the vehicle closer to the endpoint and if the boom is moving towards 95\% of the max lift.
The reward is scaled such that moving the lift correctly between two timesteps will result in the same order of magnitude as the reward of driving forward between two timesteps.
Lastly, there is a punishment based on how many timesteps have elapsed in the current episode.

\subsection{Real environment setup} \label{section:exp_setup:real_env_setup}
The real environment uses an experimental Volvo L180H wheel loader that differs quite a bit from the production version.
This vehicle is equipped with both custom software to be able to be programmatically controlled and custom hardware that facilitates this type of automation not present on production machines.
The interface, exact sensor setup, and internal code cannot be shared publicly due to it being proprietary.

To examine if the agent is capable of performing the same task in reality, the goal is defined to be the same as in simulation.
This goal was for the agent to drive forward to a pre-defined position while engaging the lift from 50\% of the max lift to 95\%.
The agent control signals are the same as in the simulator.

The system is built similarly to the simulation, such that information is pulled from the sensors to form the observation, and then the agent uses this to decide the action, which is processed such that the action is equated to some functionality on the wheel loader.
The observation is a 5-dimensional vector that includes the position of the wheel loader compared to the position of the dumping point, the velocity of the vehicle, and the current lift joint angle.
The position is in the Universal Transverse Mercator (UTM) format, meaning that the current position is defined as northings (y) and eastings (x). 

To circumvent small changes in the position of the wheel loader, all positions are calculated relative to the starting position of the system.
This allows us to validate the solution at any starting position rather than defining a single one, or having to deal with small variations in the starting position.
From the starting position, together with the orientation, $\alpha$, of the wheel loader, the stopping position is defined as 5m straight in front of the vehicle.
To obtain the heading, dual GNSS sensors were placed on the wheel loader.
This means that when the agent is engaged the position part of the observation will be $x = 5 * sin(\alpha)$ and $y = 5 * cos(\alpha)$.

A PID controller was used to control the throttle to achieve a constant velocity of 2 m/s, matching the constant velocity in the simulation.
The PID can control the throttle and the brake, independent of the agent, where the PID can engage the brake to slow down if needed.
To be able to control the velocity the PID uses the current velocity of the vehicle as the feedback.

If the agent predicts to start engaging the brake this does not equate to pressing down the brake pedal completely.
This is because that type of braking behaviour can lead to the wheel loader tipping forward, depending on the material weight in the bucket and the current boom angle.
Instead, the brake will start at some initial pedal value, and then taper off slowly, resulting in a smooth stop.

The starting position of the lift actuation will be a value close to 50\% of the maximum lift value with some variance as the boom does not instantly stop when the lift is set to zero. 
From the literature, a reasonable lift actuation value at the beginning of the approach to the tipping body is around $50\%$ \cite{FRANK20181}.

Lastly, the update rate of the real vehicle is around 10\% of the update rate in simulation.

\section{Results} \label{section:results}

\begin{figure}
    \begin{center}
        \resizebox{0.95\columnwidth}{!}{\input{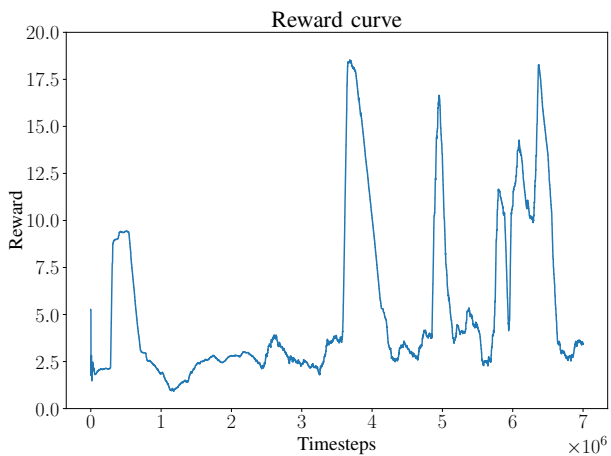}}
    \end{center}
    \caption{Reward curve for the agent trained in simulation. The agent was trained for $3\cdot10^6$ timesteps.} 
    \label{fig:results:reward_graph}
\end{figure}

The training goal is for the agent to learn how to drive forward and stop at a point while lifting the boom upwards.
The main metric used within this work is the maximum accumulated reward.
The maximum accumulated reward achieved from the entire training run was $18.5$.
Figure \ref{fig:results:reward_graph} shows the reward changes throughout the training.
From this, the training seems to collapse multiple times, however, from qualitatively validating the best-performing agent in simulation, it seems that it correctly performs the task.

\begin{figure*}
    \centering
    \includegraphics[width=\textwidth]{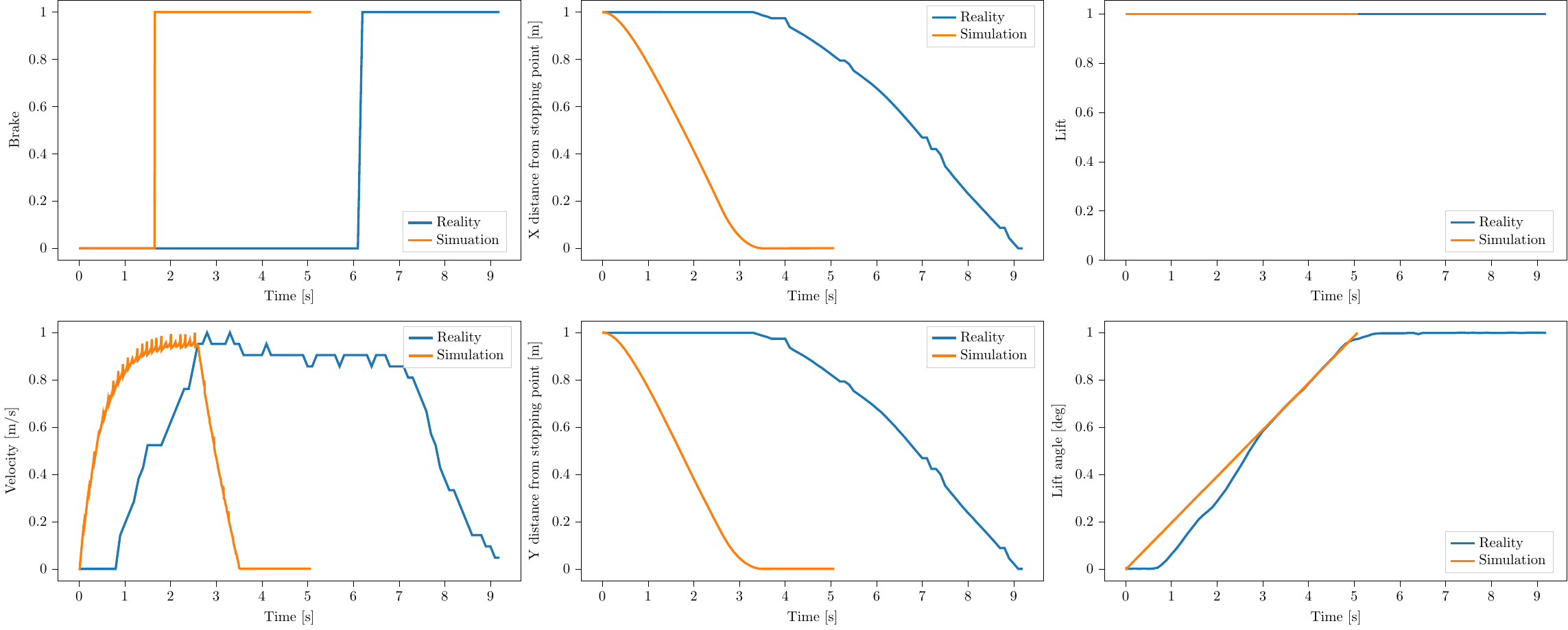}
    \caption{Results from the simulation (orange) and real vehicle (blue). The brake and lift are the predicted actions from the agent while the observation consists of velocity, distance to the stopping point in x, distance to the stopping point in y, and lift angle. All values have been normalised due to the use of a proprietary interface. For example, the top speed of the wheel loader in the simulation was 2 m/s which is normalised to 1 m/s in the figure.}
    \label{fig:result_figure}
\end{figure*}

The trained agent was transferred to a real wheel loader to examine if it was possible to train it in simulation and then validate it on a real vehicle.
Figure \ref{fig:result_figure} shows, in orange, this agent's action generated from every observation. 
As previously mentioned, the action consists of the brake and lift, while the observation consists of the x position, y position, velocity, and lift angle.
In the simulation, the agent was validated through 100 runs where the mean of the accumulated reward was 17.8, with a large variance of 16.6.
From manual tests, there appears to be some edge cases where either the initial x position or initial y position is very close to zero where the agent engages the brake the entire time while also engaging the boom.
This leads to an accumulated reward of around zero for the full episode.

This agent was then transferred to a real wheel loader with no additional training on the machine.
Figure \ref{fig:result_figure} shows the actions and observation variables in blue.
The action consists of brake and lift.
The observation consists of the x position, the y position, the velocity and the lift angle, precisely the same as in the simulation.

From Figure \ref{fig:result_figure} the agent exhibits similar behaviour on the real machine as in the simulation.
The position sensor on the real vehicle has a large delay of ~3 seconds, leading to the brake being engaged later on the real vehicle, compared to in simulation.
The consequence of this is that the real vehicle will travel further than the task.
There seems to be a difference in how fast the PID will accelerate in simulation, compared to the real vehicle.
The lift action is identical in both, together with the angular velocity of the boom being very similar.

\section{Discussion} \label{section:discussion}
The results presented in Section \ref{section:results} indicate that it is possible to use this type of development pipeline when leveraging RL to further automate the short-loading cycle.
Due to the scope of experiments, more research is needed to fully confirm this type of development pipeline.

From Figure \ref{fig:results:reward_graph}, the training collapses multiple times even when the resulting agent is capable of performing the task in both simulations and on a real wheel loader.
We hypothesise this is for two reasons. 
The first is, as mentioned, that there are some starting positions where the agent cannot perform the task, resulting in around zero accumulated reward.
This means that the data of which the agent is trained contains mostly stationary actions to get the maximum reward, leading to collapse.
The second is that because the task is quite simple to solve leading to the agent attempts to optimize more where it is not possible, leading to catastrophic forgetting.

From Figure \ref{fig:result_figure} it is apparent that the position sensors have quite a long delay, around 3 seconds, which will make it difficult to perform this type of task.
The results do indicate that once the x and y coordinates start updating, the agent does exhibit the correct behaviour.
This is achieved even with the difference in acceleration time, and differences in update rate.
The agent seems to be capable of learning the task even by using different wheel loader models in simulation and reality.
This is an encouraging sign as it does not seem that the simulation has to be extremely close to the real world for the agent to learn and perform the task.

These experiments have very limited scope such that the agent is only required to control the vehicle through two binary variables, lift and brake.
This has the consequence that the actual braking and lifting speed cannot be affected by any of the agent's choices.
The other required actions used to perform the short-loading cycle, namely throttle, steering, and tilting were either abstracted away or some other system performs said action.

This has the effect of making it impossible for the agent to learn to perform the cycle at the optimal performance.
To be able to do so, the agent has to control the wheel loader in the same way as an operator, which can affect how fast the boom moves, the velocity, and the steering.
This complicates the use of simulation that is close enough to reality for the agent to learn the task.

Introducing steering might be the most difficult part of using this type of development pipeline.
This is because the steering of the wheel loader depends on a wide set of different state variables, wheel loader mechanics, and difficult-to-model interactions, such as tyre-to-terrain.
As this can be difficult to model, the agent has to be capable of bridging the gap between simulation and reality, bypassing the need for an extremely accurate mathematical model.

As shown in Algorithm \ref{alg:experimental_setup:reward_function}, there is a term that punishes the agent for spending too long in the episode.
This was to, hopefully, make the agent learn to find a terminal state as soon as possible, meaning that the agent should minimize time.
Qualitatively no large difference could be observed when comparing the training with $t_c > 0$ compared to $t_c = 0$.

Lastly, the advantages of reinforcement learning, including the ability to learn online and learn without expert operator data, play a crucial role in automating wheel loaders.
These vehicles are highly versatile, operating in diverse environments under various conditions.
As the automation of the short-loading cycle continues to advance, transitioning away from operator-capable vehicles, these advantages gain even greater significance.
However, several uncertainties persist in the automation of the short-loading cycle, requiring further research.

\section{Conclusion} \label{section:conclusion}
The short-loading cycle is a repetitive task performed in high quantities, making it a good candidate for automation.
The wheel loader usage during the short-loading is representative of the general usage of wheel loaders in the construction industry.

In this work, the possibility of using RL for navigation during the short-loading cycle is examined.
An agent is trained to perform a simplified version of the navigation during the short-loading cycle where the operator approaches the dump truck.
The agent should learn to approach and stop at a given point while engaging the boom to lift the bucket.
The agent is trained in a simulated environment once trained, the agent is transferred to a real vehicle with no additional training.

The main contributions relate to showing that the use of RL in this domain, where an agent is trained through the use of a reward function, has the potential for the automation of the short-loading cycle.
The agent is capable of learning the task in simulation at a reasonable performance, while also being capable of performing the same task on a real vehicle with no additional training.
Due to limitations in the experimental scope together with issues regarding the sensors used on the real vehicle, more research is needed to further examine the potential of this development pipeline for automating the short-loading cycle.
To further automate this task, the agent needs to be able to predict more control signals of the wheel loader, including throttle and steering.
This will result in a larger simulation gap, which the agent will have to bridge.

Future work includes introducing more variables for the agent to control, removing abstractions between the action generated from the agent and how the vehicle itself is controlled, further shaping the reward function for productivity and energy efficiency, and attempting to use sensors that provide higher observation resolution, such as cameras.

\section*{Acknowledgement}
This research was conducted with support from Sweden's Innovation Agency and the VALD project under grant agreement no. 2021-05035. 
We want to thank Mikael Fries for his contribution to the real-world validation.

\bibliographystyle{./IEEEtran}
\bibliography{./main}

\end{document}